\documentclass[10pt,twocolumn,letterpaper]{article}

\usepackage[pagenumbers]{cvpr}
\usepackage[table,dvipsnames]{xcolor}
\usepackage{svg}
\usepackage{graphicx}
\usepackage{caption}
\usepackage{subcaption}
\usepackage{siunitx}
\usepackage{tabularx}
\usepackage{multirow} 
\usepackage{makecell} 
\usepackage{colortbl} 
\usepackage{diagbox} 

\usepackage{graphicx}
\usepackage{multirow} 
\usepackage{makecell} 
\usepackage{colortbl} 
\usepackage{diagbox} 
\usepackage{siunitx}
\usepackage{enumitem} 
\usepackage{arydshln} 
\usepackage{enumitem} 
\usepackage{arydshln} 
\usepackage{tabularx}
\usepackage{float}
\usepackage{booktabs} 
\usepackage{xspace} 
\usepackage[Export]{adjustbox} 
\usepackage{pifont} 
\usepackage{dsfont}
\usepackage[accsupp]{axessibility}  
\usepackage{bbding}

\colorlet{colorFst}{Green!25}       
\colorlet{colorSnd}{SpringGreen!45} 
\colorlet{colorTrd}{Yellow!30}      
\colorlet{colorLow}{darkgray!30}    
\newcommand{\fst}{\cellcolor{colorFst}\bf}   
\newcommand{\nd}{\cellcolor{colorSnd}}      

\newcommand{\ours}{IG-SLAM\xspace}

\usepackage[dvipsnames]{xcolor}

\definecolor{cvprblue}{rgb}{0.21,0.49,0.74}
\usepackage[pagebackref,breaklinks,colorlinks,citecolor=cvprblue]{hyperref}
\usepackage{amsmath,amsfonts}
\usepackage{algorithmic}
\usepackage{algorithm}
\usepackage{array}
\usepackage{textcomp}
\usepackage{stfloats}
\usepackage{url}
\usepackage{verbatim}
\usepackage{graphicx}
\usepackage{cite}
\usepackage{tabularx}

\newcommand{\norm}[1]{\left\lVert#1\right\rVert}

\NewDocumentEnvironment{alignb}{b}{%
  \begin{align*}
  \refstepcounter{equation} #1 \tag{\theequation}
  \end{align*}
}{\ignorespacesafterend}

\def\paperID{50} 
\def\confName{3DV\xspace}
\def\confYear{2025\xspace}

\title{IG-SLAM: Instant Gaussian SLAM}

\author{F. Aykut Sarıkamış \qquad A. Aydın Alatan\\
Center for Image Analysis (OGAM), EEE Department, METU, Turkey
}

\begin{document}
\twocolumn[{
\renewcommand\twocolumn[1][]{#1}
\maketitle
\vspace{-2.0em}
    \includegraphics[width=\textwidth]{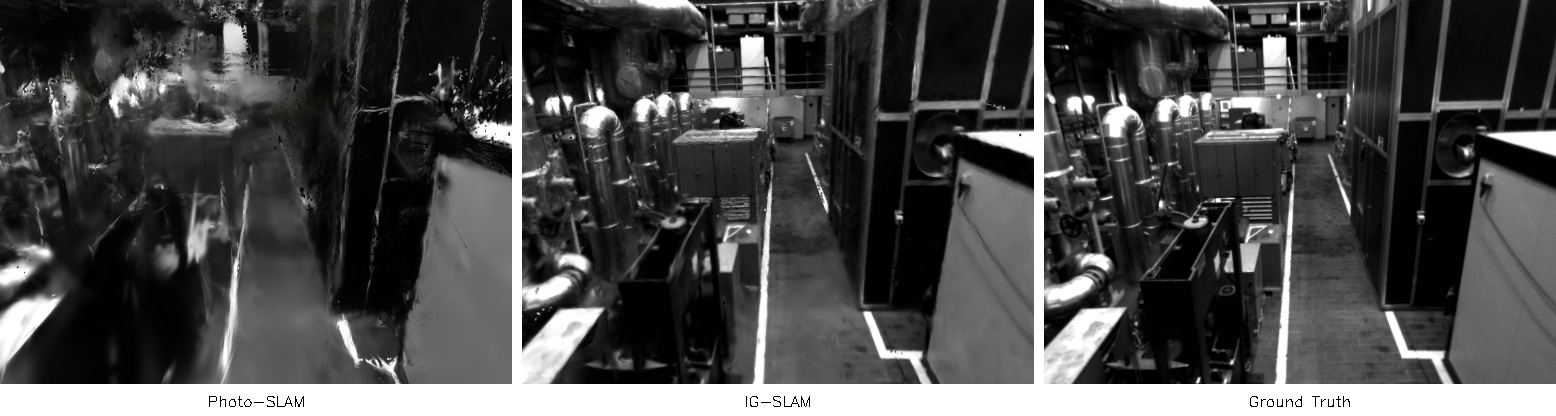}    
    \captionof{figure}{\small\textbf{Qualitative rendering results from Photo-SLAM \cite{photo-slam} and IG-SLAM.} We compare the visual quality of the methods on the large-scale EuRoC dataset \cite{euroc}.}
    \label{fig:teaser}
\vspace{0.5em}
}]

\begin{abstract}
3D Gaussian Splatting has recently shown promising results as an alternative scene representation in SLAM systems to neural implicit representations. However, current methods either lack dense depth maps to supervise the mapping process or detailed training designs that consider the scale of the environment. To address these drawbacks, we present IG-SLAM, a dense RGB-only SLAM system that employs robust dense SLAM methods for tracking and combines them with Gaussian Splatting. A 3D map of the environment is constructed using accurate pose and dense depth provided by tracking. Additionally, we utilize depth uncertainty in map optimization to improve 3D reconstruction. Our decay strategy in map optimization enhances convergence and allows the system to run at 10 fps in a single process. We demonstrate competitive performance with state-of-the-art RGB-only SLAM systems while achieving faster operation speeds. We present our experiments on the Replica, TUM-RGBD, ScanNet, and EuRoC datasets. The system achieves photo-realistic 3D reconstruction in large-scale sequences, particularly in the EuRoC dataset. 
\end{abstract}    
\section{Introduction}
\label{sec:intro}

Dense Simultaneous Localization and Mapping (SLAM) is a fundamental problem in computer vision with numerous applications in robotics, augmented reality, virtual reality, and more. Any SLAM system must operate in real-time and scale to large scenes for all these real-world applications. Additionally, the system must be robust against noisy visual sensor measurements.

The prominent scene representation is a 3D point cloud in traditional Dense SLAM systems. However, point clouds are an impoverished representation of the world. As a sparse representation, the point clouds do not provide water-tight, photo-realistic depictions of the environment. Recently, two promising scene representations have been introduced and studied in the SLAM literature: Neural Radiance Fields (NeRF) \cite{og-nerf} and Gaussian Splatting \cite{og-gs}.

Earlier dense SLAM studies that equip NeRF as an only-scene representation \cite{imap,nice-slam} achieved 3D reconstruction without camera poses in real-time. Several following studies \cite{orbeez,nerfslam,go-slam} integrate classical SLAM methods such as tracking by feature matching, dense-bundle adjustment, loop closure, and global bundle adjustment. Several performance improvements are made in later studies \cite{e-slam,co-slam,glorie,mod-slam,nicer-slam} by incorporating additional data structures along with NeRF \cite{og-nerf}, by employing off-the-shelf tracking modules \cite{Orb-slam2,droid-slam} and monocular depth estimation \cite{omnidata}. However, NeRF suffers from slow rendering speed \cite{kilonerf}; since the real-time operation is crucial for a SLAM system, slow rendering speed puts NeRF into a disadvantageous position as a scene representation. 

Later the following studies incorporate Gaussian Splatting as scene representation: Early works \cite{gaussian-splatting-slam,gaussian-slam,gs-slam} adopt Gaussian Splatting as an only-scene representation and simultaneously track and map the environment in real-time. However, utilizing novel view synthesis methods as both tracking and mapping tools is compelling yet challenging. The difficulty arises because pose and map optimizations are performed jointly. To decouple these two daunting tasks, \cite{photo-slam,splat-slam} utilize traditional SLAM methods demonstrating superior performance over only-scene representation methods in terms of reconstruction. However, these studies either lack dense depth supervision or a high frame rate.

Purposely, we introduce IG-SLAM, a deep-learning-based dense SLAM system that achieves photo-realistic 3D reconstruction in real-time. The proposed system features robust pose estimation, refined dense depth maps, and Gaussian Splatting representation. The proposed system frequently performs global dense bundle adjustment to reduce drift. Since the pose and depth maps optimized by a dense SLAM system are often noisy, we utilize depth uncertainty to make the mapping process robust to noise. Our efficient mapping algorithm is optimized specifically to work with dense depth maps enabling our system to operate at high frame rates. We perform extensive experiments on various indoor RGB sequences, demonstrating the robustness, fast operation speed, and scalability of our method.
In summary, we make the following contributions:   

\begin{itemize}
  \item We present IG-SLAM, an efficient dense RGB SLAM system that performs at high frame rates, offering scalability and robustness even in challenging conditions.
  \item A novel 3D reconstruction algorithm that accounts for depth uncertainty, making the 3D reconstruction robust to noise.
  \item A training procedure to make dense depth supervision for the mapping process as efficient as possible. 
\end{itemize}
\section{Related Work}

\subsection{Dense Visual SLAM}

Pioneering dense SLAM algorithms, DTAM \cite{DTAM} and KinectFusion \cite{KinectFusion}, show that dense SLAM can be performed in real-time despite its computational complexity. DTAM aims to produce dense depth maps associated with the keyframes, known as the view-centric approach. Later research adopted a similar approach but with a crucial distinction. While these traditional approaches generally decouple the optimization of dense maps and poses, some recent works focus on joint optimization. However, optimization of the full-resolution depth map is not feasible due to the high number of independent variables. Therefore, the following research focuses on reducing the computational complexity of joint optimization. For this purpose, BA-Net \cite{BA-Net} includes a depth map into the bundle adjustment layer utilizing a basis of depth maps and optimizing the linear combination coefficients. Code-SLAM \cite{code-slam} reduces the dimension of dense maps by an autoencoder-inspired architecture. DROID-SLAM \cite{droid-slam} optimizes down-sampled dense maps in a bundle-adjustment layer with a reprojection error, aided by optical flow revisions \cite{raft}. A recent work, FlowMap \cite{flowmap}, estimates a dense depth map with a convolutional neural network and calculates the pose analytically using the optical flow. As world-centric alternatives to this approach, Neural Radiance Fields \cite{og-nerf} and Gaussian Splatting \cite{og-gs} are utilized in the literature.  

\subsection{Neural Radiance Field Scene Representation}

NeRF \cite{og-nerf} encodes the scene as radiance fields utilizing a simple multi-layer perceptron (MLP). The original NeRF formulation exhibits slow training and rendering speeds. However, several improvements have been proposed on this initial formulation. The cone-shaped rendering \cite{mip-nerf} is utilized to address anti-aliasing, additional data structures are also employed, such as voxel grid \cite{voxelgrid1,voxelgrid2,voxelgrid3}, plenoctree \cite{plenoctree1,plenoctree2,plenoctree3}, hash tables \cite{instant-ngp} and many more achieve orders of magnitude faster rendering and training compared to the original NeRF \cite{og-nerf}. Surface-based methods \cite{neus,surface1,surface2} also unify surface and volume rendering. 

The landmark work iNeRF \cite{inerf} calculates camera poses given a NeRF representation by fixing the NeRF representation and minimizing rendering error by optimizing the camera pose around an initial guess. iMAP \cite{imap}, as the first representation-only work, optimizes the pose by fixing the NeRF representation and optimizes the map based on the calculated pose. NICE-SLAM \cite{nice-slam} introduces a hierarchical coarse-to-fine mapping approach. To decouple map and pose optimization, Orbeez-SLAM \cite{orbeez} leverages robust visual SLAM methods \cite{Orb-slam2} and multi-resolution hash encoding \cite{instant-ngp}. NeRF-SLAM \cite{nerfslam} introduces dense depth maps with covariance and poses generated by the robust dense-SLAM algorithm DROID-SLAM \cite{droid-slam}. GO-SLAM employs loop closing and global dense bundle adjustment to achieve globally consistent reconstruction. NICER-SLAM \cite{nicer-slam} extends NICE-SLAM \cite{nice-slam}  incorporating off-the-shelf monocular depth and normal estimators. Recently, MoD-SLAM \cite{mod-slam} utilizes cone-shaped projection in rendering \cite{mip-nerf}. GlORIE-SLAM \cite{glorie} utilizes monocular depth estimation for mapping supervision.

\subsection{3D Gaussian Splatting Scene Representation}
3D Gaussian Splatting represents the scene as a set of Gaussians of varying colors, shapes, and opacity. Several improvements are proposed for consistency and reconstruction quality. For example, 2D counterpart \cite{2d-gs} is also proposed to enhance multi-view consistency. Moreover, the rendering depth with alpha-blending as in the original 3D Gaussian Splatting causes noisy surfaces; hence, more rigorous methods address this issue by utilizing varying depths per Gaussian according to the viewpoint \cite{pgsr,rade-gs}.

Due to its fast rendering speed and being an explicit scene representation as opposed to NeRF \cite{og-nerf}, Gaussian Splatting \cite{og-gs} has also quickly gained attention in the SLAM literature. MonoGS \cite{gaussian-splatting-slam}, GS-SLAM \cite{gs-slam}, and SplaTAM \cite{splatam} are pioneering Gaussian-Splatting representation-only SLAM algorithms that jointly optimize Gaussians and the pose. Gaussian-SLAM \cite{gaussian-slam} introduces sub-maps to mitigate neural forgetting. Photo-SLAM \cite{photo-slam} decouples tracking and mapping by employing a traditional visual SLAM algorithm \cite{Orb-slam2} as its tracking module and introduces a coarse-to-fine map optimization approach. RTG-SLAM \cite{rtg-slam} renders depth by considering only the foremost opaque Gaussians. Recent work, Splat-SLAM \cite{splat-slam} uses proxy depth maps to supervise map optimization. 

\section{Proposed Method}

\begin{figure*}[t!] 
  \centering
  \includegraphics[width=\textwidth]{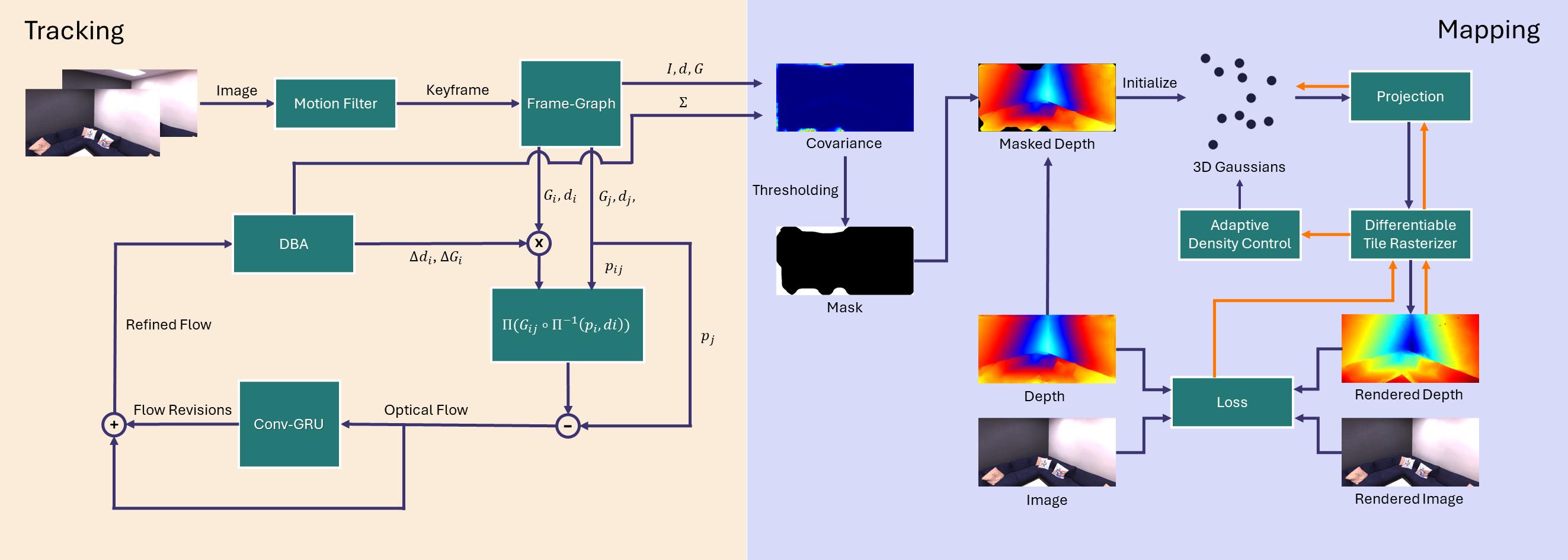} 
  \captionsetup{justification=justified}
  \caption{\textbf{System Overview.} Our system takes an RGB image stream as input and outputs the camera pose and scene representation in the form of a set of Gaussians. We decouple this objective into two parts: tracking and mapping. \textbf{Tracking:} Keyframes are created and added to the frame graph based on average optical flow. Pretrained GRU refines optical flow between keyframes. Dense bundle adjustment (DBA) is performed on the frame graph, minimizing reprojection error while optimizing the dense depth map and camera pose, and calculating depth map covariance simultaneously. After several iterations, depth maps and camera poses are expected to converge. \textbf{Mapping:} Keyframes' pose, depth, and covariance obtained from tracking are used for 3D reconstruction. We initialize Gaussians from low covariance regions utilizing the camera pose and depth map. 3D Gaussians are then projected onto the image plane and rendered utilizing a differentiable tile rasterizer. The loss function is a combination of depth and color loss. The depth loss is weighted by covariance. Finally, the loss is backpropagated to optimize Gaussians orientation, scaling, opacity, position, and color designated by orange arrows in the figure. Moreover, Gaussians are split, cloned, and pruned based on the local gradients. }
  \label{fig:overviewofsystem}
\end{figure*}

We provide an overview of the proposed method in \cref{fig:overviewofsystem}. Our tracking algorithm (\cref{subsec:Tracking}) generates a dense depth map, depth uncertainty, and the camera pose for each keyframe. These outputs are then used to supervise our mapping algorithm (\cref{subsec:Mapping}). The Gaussians are initialized based on the camera pose and dense depth and are optimized using color and weighted depth loss. Real-time operation is achieved through a sliding window of keyframes.

\subsection{Tracking}
\label{subsec:Tracking}

We mainly employ DROID-SLAM \cite{droid-slam} as our tracking module. DROID-SLAM maintains two state variables: camera pose $\mathbf{G}_t$ and inverse depth $\mathbf{d}_t$ for each camera frame $t$. DROID-SLAM constructs a frame graph $(\mathcal{V}, \mathcal{E})$ of keyframes based on co-visibility. Keyframes are selected from all camera frames when the average magnitude of the optical flow for a frame is higher than a certain threshold. If there is a visual overlap between frames $i$ and frame $j$, an edge is created between the $i^{th}$ and $j^{th}$ vertex in $\mathcal{V}$. This graph is updated during inference. Given the initial pose and depth estimates $(\mathbf{G}_i$,$\mathbf{d}_i$) and $(\mathbf{G}_j$,$\mathbf{d}_j$) for frame $i$ and $j$, the optical flow field is estimated by unprojecting the pixels from frame $i$, projecting them into frame $j$, and taking the pixel-wise position difference. In other words, the reprojected pixel locations $p_{ij}$ is calculated as in \cref{eqn:projection}
\begin{equation}
\label{eqn:projection}
p_{ij} = \Pi(\mathbf{G}_{ij} \circ \Pi^{-1}(\mathbf{p}_{i},\mathbf{d}_i)),\;\; \mathbf{p}_{ij} \in \mathbb{R}^{H \times W \times 2}
\end{equation}
where $\mathbf{G}_{ij} = \mathbf{G}_{j}^{-1} \circ \mathbf{G}_{i}$. Then, the optical flow is initially calculated as $p_{ij} - p_{j}$. This estimate is fed into GRU along with a correlation vector which is an inner product between features of the frames. The GRU produces flow revisions $\mathbf{r}_{ij}$ and confidence weights $\mathbf{w}_{ij}$. the refined reprojected pixel locations $\mathbf{p}^{\ast}_{ij}$ are computed similarly to \cref{eqn:projection}  incorporating the flow correction from the GRU. Then, the dense bundle adjustment layer minimizes the cost function in \cref{eqn:error}.

\begin{alignb}
\label{eqn:error}
&\mathbf{E}(\mathbf{G'},\mathbf{d'}) = \sum_{i,j \in \mathcal{E}}\norm{\mathbf{p}^{\ast}_{ij} - \mathbf{p'}_{ij}}^2_{\Sigma_{ij}} \\
&\mathbf{p'}_{ij} = \Pi(\mathbf{G'}_{ij} \circ \Pi^{-1}(\mathbf{p}_{i},\mathbf{d'}_i))
\end{alignb}
where $\Sigma_{ij} = $ diag$(\mathbf{w}_{ij})$ and $\norm{\mathbf{.}}_\Sigma$ Mahalanobis norm weighted according to the weights $\mathbf{w}_{ij}$. Linearizing \cref{eqn:error} around $(\mathbf{G'},\mathbf{d'})$ and solve for pose and depth updates $(\Delta \mathbf{\xi},\Delta \mathbf{d})$ using Gauss-Newton algorithm. The linearized system of equations becomes 

\begin{equation}
\label{eqn:hessian}
H\textbf{x} = \textbf{b}, \;\; H = \begin{bmatrix}
    C & E \\
    E^T & P
\end{bmatrix}, \;\; \mathbf{x} = \begin{bmatrix}
   \Delta \mathbf{\xi} \\
   \Delta \mathbf{d}
\end{bmatrix}, \mathbf{b} = \begin{bmatrix}
   \mathbf{v} \\
   \mathbf{w}
\end{bmatrix}
\end{equation}
where $H$ is the Hessian matrix, $\mathbf{x} = [\Delta \mathbf{\xi}, \Delta \mathbf{d}]$ is the pose and depth updates, $\mathbf{b}=[\mathbf{v}, \mathbf{w}]$ is the pose and depth residuals, $C$ is the block camera matrix. $E$ is the camera/depth off-diagonal block matrices, and $P$ is the diagonal matrix corresponding to disparities per pixel per keyframe. The bundle adjustment layer operates on the initial flow estimates and updates the keyframes' pose and depth map. Optical flow is then recalculated by refined poses and depth maps which are subsequently fed back into the dense bundle adjustment layer. After successive iterative refinements on the keyframe graph, the poses and depth maps are expected to converge. 

After the dense bundle adjustment step, we compute the covariance for depth estimates. As shown in NeRF-SLAM \cite{nerfslam}, the same Hessian structure in \cref{eqn:hessian} can be used to calculate covariance for depth estimates $\Sigma_d$ and poses $\Sigma_G$ as shown in \cref{eqn:cov}. The depth covariance is used both as a mask for initializing Gaussians and as weights in the depth component of the loss function.

\begin{alignb}
&\Sigma_d = P^{-1} + P^{-T}E^T\Sigma_{G}EP^{-1} \\
&\Sigma_G = (LL^T)^{-1}
\label{eqn:cov}
\end{alignb}

\noindent \textbf{Keyframing} We utilize all the keyframes that are actively optimized in the tracking process without any filtering. Each keyframe that participates in mapping contains its camera image $I$, depth map $\mathbf{d}$, depth covariance $\Sigma_d$, and pose $\mathbf{G}$. The mapping process accepts a keyframe only if it is not already in the sliding window. Note that, we do not send all the keyframes created in a mapping cycle, but only the most recent one. Therefore, this approach may result in some keyframes being missed during optimization. However, this design choice prevents abrupt changes in the sliding window caused by a sharp camera movement. \\

\noindent \textbf{Global BA} After the number of total keyframes exceeds the sliding window length for the Dense Bundle Adjustment, we regularly perform Global Bundle Adjustment for all existing keyframes on a separate graph as described in GO-SLAM \cite{go-slam}. The graph is constructed utilizing a distance metric, where the distance between frame pairs is the average optical flow magnitude. Graph edges are established between consecutive keyframes and those that are close according to the distance metric. Dense bundle adjustment is then applied based on this graph every 10 keyframes. The pose and depth maps are updated at the start of every mapping cycle, along with their covariances. We perform one last global BA at the end of tracking.

\subsection{Mapping}
\label{subsec:Mapping}

The mapping process is responsible for 3D reconstruction with keyframes equipped with pose, image, depth, and covariance acquired from the tracking process. \\

\noindent \textbf{Representation} We adopt Gaussian Splatting \cite{og-gs} as scene representation. A Gaussian function is described by \cref{eqn:gaussianeqn}
\begin{equation}
\label{eqn:gaussianeqn}
G(\mathbf{x}) = \exp\biggl(\frac{1}{2}(\mathbf{x}-\mathbf{\mu})^T\Sigma^{-1}(\mathbf{x}-\mathbf{\mu})\biggr)
\end{equation}
where $\mathbf{\mu}$ and $\Sigma$ are the mean and covariance which define the position and shape of this Gaussian. To ensure that the covariance remains semi-definite during optimization, covariance $\Sigma$ is decomposed into $RSS^TR^T$ where $R$ is the rotation matrix and $S$ is the scaling matrix. In addition to position, rotation, and scaling, opacity $\alpha$ and color $c$ are also optimized. Although the original implementation parameterizes color as spherical harmonic coefficients, our algorithm optimizes the color directly. The projection of a 3D covariance is formulated as $\Sigma' = JR\Sigma R^TJ^T$ where $R$ is the rotation component of the world-to-camera transformation $T_{cw}$  and $J$ is the Jacobian of the affine approximation of the projective
transformation $P$ \cite{zwicker2001ewa}. The position is projected directly as $\mu' = PT_{cw}\mu$. \\

\noindent \textbf{Rendering} A set of Gaussians $\mathcal{N}$ visible from a viewpoint, is first projected onto the image plane. 2D Gaussians are then sorted according to their depths and are rasterized via $\alpha$-blending as described in \cref{eqn:alpha-blending} for color and depth.
 
\begin{equation}
\label{eqn:alpha-blending}
\hat{C} = \sum_{i \in \mathcal{N}}c_i\alpha_i\prod_{j=1}^{i-1}(1-\alpha_j), \; \hat{D} = \sum_{i \in \mathcal{N}}d_i\alpha_i\prod_{j=1}^{i-1}(1-\alpha_j)
\end{equation}

\noindent \textbf{Hierarchical Optimization} Since dense depth maps for keyframes are available, we adopt a training strategy similar to RGB-D MonoGS \cite{gaussian-splatting-slam} but utilizing a coarse-to-fine training strategy inspired by Photo-SLAM \cite{photo-slam} and Instant-NGP \cite{instant-ngp}. 

For each keyframe, an image pyramid is constructed by downsampling image, depth, and covariance by a factor of $s$ using bilinear interpolation, as in \cref{eqn:keyframepyr}

\begin{alignb}
\label{eqn:keyframepyr}
&\text{KF}_i^l = \{I_i^l, \mathbf{d}_i^l, \Sigma_{di}^l\} \\
&I_i^l = I_i^0 \downarrow s^l, \;\; \mathbf{d}_i^l = \mathbf{d}_i^0 \downarrow s^l, \;\; \Sigma_{di}^l =\Sigma_{di}^0 \downarrow s^l
\end{alignb}
where $\downarrow$ denotes the downsampling operation with linear interpolation and $l$ is the pyramid level and $I_i^0$, $\mathbf{d}_i^0, \Sigma_{di}^0$ are the full resolution image, depth, and covariance respectively. In Photo-SLAM \cite{photo-slam}, the authors utilize a sharp downsampling factor $s$ of 0.5 and a 2-level pyramid. In contrast, we employ a smoother downsampling factor $s=0.8$ similar to Instant-NGP \cite{instant-ngp} and a 3-level pyramid. 

\begin{figure}[h]
    \centering
    \includegraphics[width=\linewidth]{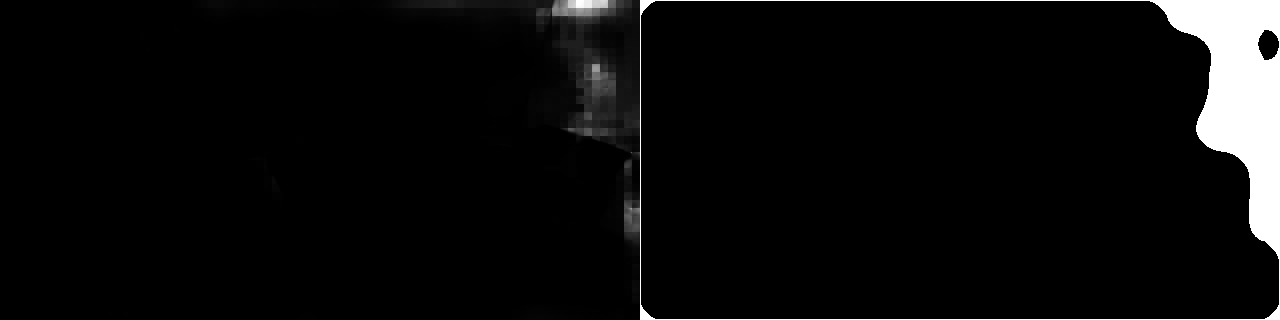}
    \caption{\textbf{An example of normalized covariance(left) and corresponding mask(right).} The mask is created by thresholding normalized covariance with a maximum filter and smoothing with a majority filter. The white region on the mask is left out and not used during Gaussian initialization.}
    \label{fig:mask}
\end{figure}

In each pyramid level, Gaussians are initialized by unprojection as follows: The points are sampled randomly from the most recent keyframe by using a downsampling factor $\theta$. The sampled points are then unprojected according to depth maps. To account for the noise in depth maps, regions with high covariance are masked out to make the Gaussian initialization more robust to noise. \cref{eqn:mask} describes a mask for a given normalized depth covariance. 

\begin{equation}
\label{eqn:mask}
M = \{(i,j) \; | \; \sigma_{ij} < 0.2\}
\end{equation}

\noindent where $M$ represents  the binary mask matrix and $i$ and $j$ represent pixel location. The mask is created by normalizing the covariance $ \Sigma$ between 0 and 1 and identifying the pixel values below 0.2 normalized covariance $\sigma$. The mask is then smoothed using thresholding operation as described in \cref{eqn:mask}  with a maximum filter followed by a majority filter. An example of a mask for a given covariance is shown in figure \cref{fig:mask}.

The map optimization is performed on a sliding window in a coarse-to-fine fashion. We maintain the last $N$ keyframes within the sliding window to meet the real-time requirements. As the number of iterations increases, we switch to training with higher resolutions in the image pyramid. At the beginning of optimization at each level $l$, Gaussians are unprotected according to its depth map $\mathbf{d}_i^l$. We render the Gaussians from keyframes' viewpoints in the sliding window, and the loss function is calculated based on the rendered image and depth. Camera images and dense depth maps are utilized as ground truth in mapping supervision. We employ a loss function that combines weighted depth loss $L_\text{depth}$ and color loss $L_\text{color}$ which are defined as below

\begin{equation}
\label{eqn:depthloss}
L_\text{depth}= \norm{D - \hat{D}}_{\Sigma_d^{-1}}^{1} , \;\;  L_\text{color} = \norm{C - \hat{C}}^{1}
\end{equation}
where $D$ and $C$ are the ground truth depth and image, respectively, and $\hat{D}$ and $\hat{C}$ are the rendered depth and image according to \cref{eqn:alpha-blending}. The depth loss $L_\text{depth}$ is weighted by the inverse covariance to ensure that the pixels with high uncertainty are weighted less. The combined loss is given by $L = \alpha L_\text{color} + (1-\alpha)L_\text{depth}$. We set $\alpha = 0.5$ throughout all of our experiments. The loss is then backpropagated through a differentiable rendering pipeline where the position, opacity, covariance, maps, and color of the Gaussians are optimized. \\

\noindent \textbf{Post Processing} We refine the mapping results by optimizing the map for several iterations following the conventions established in MonoGS \cite{gaussian-splatting-slam},  GlORIE-SLAM \cite{glorie} and Splat-SLAM \cite{splat-slam}. For this purpose, we randomly select single frames and optimize the map with the same loss function used in the mapping. We perform the same number of iterations in MonoGS \cite{gaussian-splatting-slam} and Splat-SLAM \cite{splat-slam} for fairness.

\subsection{Training Strategy}

\begin{figure}[!ht]
    \centering
    \includegraphics[width=\linewidth]{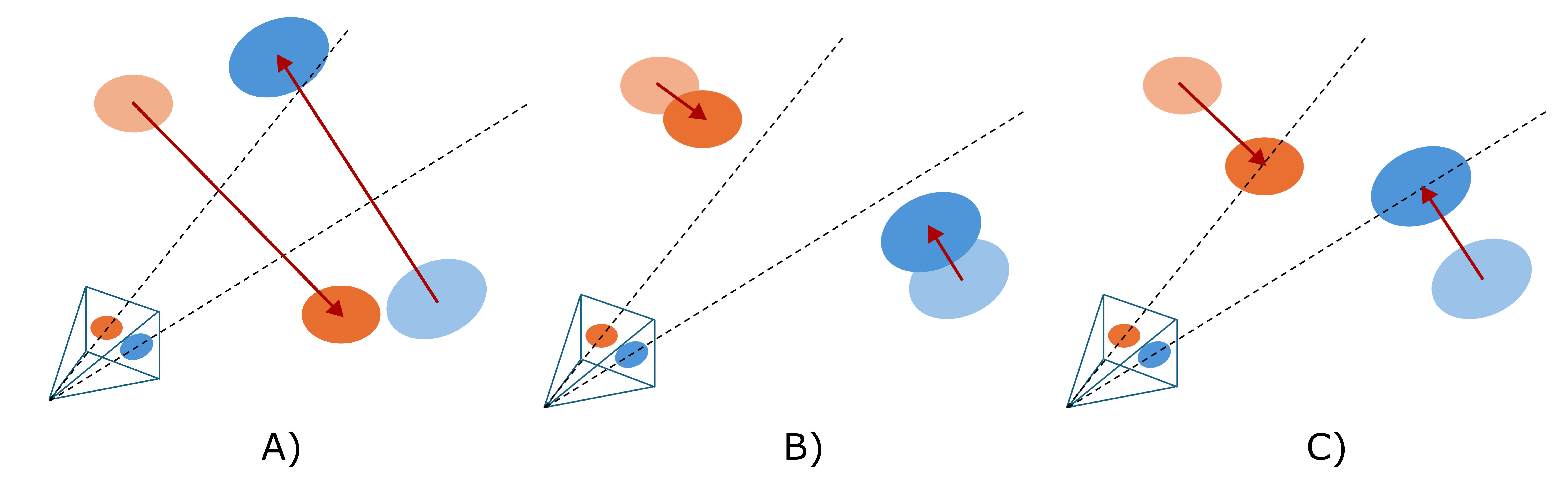}
    \caption{\textbf{Three hypothetical cases to encounter in training.} Dashed lines pass through ground truth Gaussian positions from the camera center. The faded Gaussians represent their previous positions. Red lines are the position update steps along the gradient direction. In \textbf{A)}, a large position update causes the order of Gaussians to change, creating TV-static-like noise in training. In \textbf{B)},  multiple iterations are needed to move Gaussians to the correct place because of small position updates. \textbf{C)} represents the ideal case where position update is exactly the position error.}
    \label{fig:TV-static}
\end{figure}

A subtle yet crucial point regarding our training strategy is that dense depth maps may be noisy; however, they are unlikely to disrupt depth order. In other words, having a position learning rate such that Gaussians switch positions during training is redundant and hinders optimization convergence. This effect is illustrated in \cref{fig:TV-static}. It should be noted that this is never the case for standard Gaussian Splatting training where the method typically starts with a sparse SfM point cloud. However, since Gaussians are initialized from a dense depth map, they are quite close to each other.

 As illustrated in \cref{fig:TV-static}, case \textbf{A)} high learning rates cause the optimization to bounce Gaussians around the desired position. Conversely, the polar opposite in \textbf{C)} also hinders the convergence. Since setting a perfect learning rate for each iteration is neither feasible nor practical, we choose a learning rate that decays during training according to \cref{eqn:decay} to reduce this TV static noise during training. We initialize the learning rate to cover the full range needed to detail the model from coarse to fine while allowing for gradual decay.
 
 \begin{equation}
\label{eqn:decay}
\text{lr}(t) = \exp((1-t)\ln(\text{lr}_i) + t \ln(\text{lr}_f))
\end{equation}

\noindent where $t = n/\tau$ is the iteration number $n$ over decay constant $\tau$, and $\text{lr}_i$, $\text{lr}_f$ are the initial and final learning rate, respectively. The impact of learning rate and its decay in training performance are examined in \cref{ch:experiments}.

We densify Gaussians in high loss gradient regions at every 150 iterations. Densification is achieved by cloning small Gaussians and by splitting large ones. The occluded Gaussians are also pruned at the end of each sliding window optimization to ensure that only the necessary Gaussians for accurate reconstruction are retained.

\section{Experiments}
\label{ch:experiments}
We evaluate our system on various synthetic and real-world datasets. The ablation studies and hyperparameter analyses are also demonstrated to justify our design choices.

\subsection{Experimental Setup}

\textbf{Datasets} We evaluate the system  in Replica \cite{Replica}, TUM RGB-D \cite{tumrgbd}, ScanNet \cite{scannet}, and EuRoC MAV \cite{euroc} datasets. Replica is a dataset of synthetic indoor scenes. The TUM RGB-D dataset consists of sequences that are recorded in small indoor office environments. The ScanNet dataset consists of 6 sequences of real-world indoor environments. The EuRoC is a dataset collected on board a Micro Aerial Vehicle (MAV) containing stereo images of relatively large-scale indoor environments. All datasets are evaluated without clipping except EuRoC. We clip from the start of the sequences to skip typical pauses at the beginning. We run all sequences 3 times and report the average results to mitigate the effect of the non-deterministic nature of multi-processing.  \\

\noindent \textbf{Metrics} Following the view synthesis SLAM literature convention, we evaluate our system using PSNR, SSIM, and LPIPS \cite{LPIPS}. We also provide depth L1[cm]  metric compared to the ground truth depth in the Replica dataset. The evaluation is performed after post-processing every 5 frames in sequences skipping the keyframes used for mapping. This approach aligns with the evaluation methods used in MonoGS \cite{gaussian-splatting-slam} and Splat-SLAM \cite{splat-slam}. \\

\noindent \textbf{Implementation Details} Our system runs on a PC with a 3.6GHz AMD Ryzen Threadripper PRO 5975WX and an NVIDIA RTX 4090 GPU. In all our experiments, we set $l=0.8$, $\theta = 128$, $\alpha = 0.5$, $\text{lr}_i = \num{1.6e-4}$, $\text{lr}_f = \num{1.6e-6}$, $\tau = 3000$ for hyperparameters in mapping. We set $\beta = 2000$ for the EuRoC \cite{euroc} and Replica \cite{Replica} datasets and $\beta = 26000$ for the TUM RGB-D \cite{tumrgbd} and the ScanNet \cite{scannet} datasets. These values are consistent with those used in MonoGS \cite{gaussian-slam} and Splat-SLAM \cite{splat-slam}. For tracking, pre-trained GRU weights from DROID-SLAM \cite{droid-slam} are utilized. We set the mean optical flow threshold for keyframe selection to 4.0 pixels, and the local dense bundle adjustment window to 16. Optimizations in the tracking module are performed in LieTorch \cite{lietorch} framework. The mapping process accepts only the latest keyframe created after finishing its optimization step if the latest keyframe is not already in the sliding window. \\

\noindent \textbf{Baselines} We compare our system to state-of-the-art RGB-only Gaussian Splatting and NeRF SLAM algorithms, including MonoGS \cite{gaussian-splatting-slam}, Photo-SLAM \cite{photo-slam}, GlORIE-SLAM \cite{glorie}, and Splat-SLAM \cite{splat-slam}.

MonoGS \cite{gaussian-splatting-slam} is the state-of-the-art representation-only SLAM algorithm that utilizes the Gaussian scene representation for tracking and mapping. Photo-SLAM, like GlORIE-SLAM \cite{glorie}, Splat-SLAM \cite{splat-slam}, and our system, features a decoupled design for tracking and mapping. One key difference is that Photo-SLAM lacks dense depth maps while mapping. GlORIE-SLAM and Splat-SLAM utilize monocular depth estimation \cite{omnidata} and the dense bundle adjustment layer. The most important difference between them is that GlORIE-SLAM \cite{glorie} models the scene with NeRF \cite{og-nerf} and Splat-SLAM \cite{splat-slam} does so with 3D Gaussian Splatting \cite{og-gs}.

\subsection{Evaluation}
We compare our system with state-of-the-art algorithms based on rendering quality, 3D reconstruction accuracy, and runtime performance. \\

\noindent \textbf{Rendering and Reconstruction Accuracy} We evaluate rendering and reconstruction accuracy for the Replica \cite{Replica} in \cref{tab:render_replica}. Our algorithm's performance is quite similar to Splat-SLAM \cite{splat-slam} in Replica \cite{Replica}. In \cref{tab:render_scannet}, we compare head-to-head with GlORIE-SLAM \cite{glorie} on the ScanNet \cite{scannet}, where we trail behind Splat-SLAM \cite{splat-slam}. In \cref{tab:render_tum}, we rank just behind Splat-SLAM, outperforming other algorithms on the TUM RGB-D \cite{tumrgbd}  dataset. However, we are superior in terms of on-the-fly map optimization to Splat-SLAM as shown in \cref{tab:office0_refinement}. We place the first in the the EuRoC \cite{euroc} dataset demonstrating a significant margin over Photo-SLAM \cite{photo-slam}. A qualitative comparison is shown in \cref{fig:teaser}. Our experiments reveal that sequences focusing on a centered object in an unbounded scene, such as TUM-RGBD \texttt{f3/off}, are particularly challenging.  \\

\begin{table}[h!]
  \footnotesize
  \setlength{\tabcolsep}{0.7em}
  \resizebox{\linewidth}{!}{
    \setlength{\tabcolsep}{8.0pt}
    \begin{tabular}{lcccccc}
    \toprule
    Metrics  & \makecell[c]{Mono-\\ GS~\cite{gaussian-splatting-slam}} & \makecell[c]{GlORIE-\\SLAM~\cite{glorie}} & \makecell[c]{Photo-\\SLAM~\cite{photo-slam}} & \makecell[c]{Splat - \\SLAM~\cite{splat-slam}} & \makebox[0.13\linewidth]{\textbf{Ours}}\\
    \midrule
    PSNR$\uparrow$    & 31.22   & 31.04 & 33.30 &\fst36.45 & \nd 36.21\\
    SSIM $\uparrow$   &  0.91 & 0.91  & 0.93  & \nd 0.95 & \fst 0.96 \\
    LPIPS$\downarrow$ &  0.21  & 0.12  & -     & \nd 0.06    &\fst 0.05 \\  \midrule
    \makecell[l]{Depth L1}$\downarrow$& - & - & -  & \fst 2.41 & \nd 4.34\\  
    \bottomrule
    \end{tabular}}
    \caption{
    \textbf{Rendering and Tracking Results on Replica~\cite{Replica} for RGB-Methods.} The results are averaged over 8 scenes and each scene result is the average of 3 runs. We take the numbers from \cite{splat-slam} except for ours. The best results are highlighted as \colorbox{colorFst}{\bf first}, \colorbox{colorSnd}{second}. Our method shows similar performance to Splat-SLAM \cite{splat-slam} and outperforms all the other methods. 
    }
    \label{tab:render_replica}
\end{table}

 \begin{table}[h!]
\centering
\footnotesize
\setlength{\tabcolsep}{0.7em}
\resizebox{\linewidth}{!}{
\begin{tabular}{llccccccc}
\toprule
Method & Metric & \texttt{0000} & \texttt{0059} & \texttt{0106} & \texttt{0169} & \texttt{0181} & \texttt{0207} & Avg.\\
\midrule

\multirow{3}{*}{\makecell[l]{MonoGS~\cite{gaussian-splatting-slam}}} 
& PSNR$\uparrow$ &16.91 & 19.15 & 18.57 & 20.21 & 19.51 & 18.37 & 18.79 \\
& SSIM $\uparrow$ & 0.62 & 0.69 & 0.74 & 0.74 & 0.75 & 0.70 & 0.71\\
& LPIPS$\downarrow$ &0.70 & 0.51 & 0.55 & 0.54 & 0.63 & 0.58 & 0.59 \\
[0.8pt] \hdashline \noalign{\vskip 1pt}

\multirow{3}{*}{\makecell[l]{GlORIE-\\SLAM~\cite{glorie}}} 
& PSNR$\uparrow$ &23.42 &\nd 20.66 & 20.41 & 25.23 & 21.28 & 23.68 & 22.45\\
& SSIM $\uparrow$ &  \fst 0.87 &  \fst0.87 & \nd 0.83 & \nd 0.84 & \fst 0.91 & 0.76 & \fst 0.85 \\
& LPIPS$\downarrow$ & \nd 0.26 & \nd 0.31 & 0.31 & 0.21& 0.44 &  0.29 & 0.30\\
[0.8pt] \hdashline \noalign{\vskip 1pt}

\multirow{3}{*}{\makecell[l]{Splat-\\ SLAM \cite{splat-slam}}} 
    &PSNR$\uparrow$& \fst 28.68&  \fst 27.69&\fst  27.70&\fst  31.14& \fst 31.15&\fst  30.49&  \fst29.48\\
     &SSIM $\uparrow$&\nd 0.83& \fst0.87 & \fst0.86& \fst 0.87& \nd 0.84& \fst 0.84& \fst 0.85\\
     &LPIPS $\downarrow$& \fst 0.19& \fst 0.15& \fst 0.18& \fst 0.15& \fst 0.23& \fst 0.19& \fst 0.18\\
[0.8pt] \hdashline \noalign{\vskip 1pt}

\multirow{3}{*}{\makecell[l]{\textbf{\ours}\\ \textbf{(Ours)}}} 
    &PSNR$\uparrow$&\nd 24.68&  20.09& \nd 25.30& \nd 27.85& \nd 25.80& \nd 26.69& \nd 25.07\\
     &SSIM $\uparrow$& 0.74& 0.68& 0.83&  0.82& 0.83&  \nd 0.78& 0.78\\
     &LPIPS $\downarrow$&  0.29&  0.39& \nd 0.22& \nd  0.19& \nd 0.27& \nd 0.27&  \nd 0.27\\
\bottomrule
\end{tabular}}

\caption{\textbf{Rendering Performance on ScanNet~\cite{scannet}.} Each scene result is the average of 3 runs. We take the numbers
from \cite{splat-slam} except for ours. Our method shows competitive performance to the state-of-the-art methods exhibiting the second high visual quality results.}
\label{tab:render_scannet}
\end{table}

 \noindent \textbf{Runtime Analysis}  We assess real-time performance of our algorithm in \cref{tab:memoryandruntime}. We benchmark the runtime on a 3.6GHz AMD Ryzen Threadripper PRO 5975WX  and an NVIDIA GeForce RTX 4090 with 24 GB of memory. Our system operates at 9.94 fps, making it 8 times faster than Splat-SLAM \cite{splat-slam} in a single-process implementation. Our method outperforms other algorithms without compromising visual quality. The reference multi-process implementation of our method achieves a frame rate of 16 fps. Our method's peak memory consumption and map size are comparable to existing methods.

 \begin{table}[h!]
\centering
\footnotesize
\setlength{\tabcolsep}{0.7em}
\resizebox{\linewidth}{!}{
\begin{tabular}{llcccccc}
\toprule
Method & Metric  & \texttt{f1/desk} & \texttt{f2/xyz} & \texttt{f3/off}  & \textbf{Avg.} \\
\midrule

\multirow{3}{*}{\makecell[l]{Photo-SLAM~\cite{photo-slam}}} 
    &PSNR$\uparrow$& 20.97& 21.07&  19.59& 20.54\\
     &SSIM $\uparrow$& 0.74& 0.73& 0.69& 0.72\\
     &LPIPS $\downarrow$&  0.23&  0.17&  0.24& 0.21\\
[0.8pt] \hdashline \noalign{\vskip 1pt}

\multirow{3}{*}{\makecell[l]{MonoGS~\cite{gaussian-splatting-slam}}} 
    &PSNR$\uparrow$& 19.67 &  16.17 &  20.63 & 18.82\\
     &SSIM $\uparrow$& 0.73 & 0.72 & 0.77 &  0.74\\
     &LPIPS $\downarrow$& 0.33 & 0.31 & 0.34 & 0.33\\
[0.8pt] \hdashline \noalign{\vskip 1pt}

\multirow{3}{*}{\makecell[l]{GlORIE-\\SLAM~\cite{glorie}}} 
    &PSNR$\uparrow$& 20.26&   25.62&  21.21 & 22.36\\
     &SSIM $\uparrow$& 0.79 &0.72 &0.72 & 0.74\\
     &LPIPS $\downarrow$&   0.31& \nd 0.09&0.32 & 0.24 \\
[0.8pt] \hdashline \noalign{\vskip 1pt}

\multirow{3}{*}{\makecell[l]{Splat-\\SLAM~\cite{splat-slam}}}
    &PSNR$\uparrow$& \fst 25.61& \fst 29.53& \fst 26.05& \fst 27.06\\
     &SSIM $\uparrow$& \fst 0.84& \fst 0.90& \fst 0.84& \fst 0.86\\
     &LPIPS $\downarrow$&  \fst 0.18&\fst 0.08&  0.20&\fst 0.15\\
[0.8pt] \hdashline \noalign{\vskip 1pt}

\multirow{3}{*}{\makecell[l]{\ours \\ \textbf{(Ours)}}}
    &PSNR$\uparrow$&\nd 24.45& \nd 26.35&  \nd 25.27& \nd 25.36\\
     &SSIM $\uparrow$&\nd 0.80& \nd 0.85& \nd 0.83&\nd 0.83\\
     &LPIPS $\downarrow$&\nd 0.20& 0.10& \fst 0.17& \nd 0.16\\
\bottomrule
\end{tabular}}
\caption{\textbf{Rendering Performance on TUM-RGBD \cite{tumrgbd}.} Each scene result is the average of 3 runs. We take the numbers
from \cite{splat-slam} except for ours. Our method demonstrates similar performance to Splat-SLAM \cite{splat-slam} in challenging indoor environments showing a clear performance margin to the other methods.}
\label{tab:render_tum}
\end{table}

\begin{table}[h!]
\centering
\footnotesize
\setlength{\tabcolsep}{0.7em}
\resizebox{\linewidth}{!}{
\begin{tabular}{llcccccc}
\toprule
Method & Metric  & \texttt{MH-01} & \texttt{MH-02} & \texttt{V1-01} &\texttt{V2-01} & \textbf{Avg.} \\
\midrule

\multirow{3}{*}{\makecell[l]{Photo-SLAM~\cite{photo-slam}}} 
    &PSNR$\uparrow$& 13.95& 14.20&  17.07& 15.68 & 15.23\\
     &SSIM $\uparrow$& 0.42& 0.43& 0.62 & 0.62 & 0.52\\
     &LPIPS $\downarrow$&  0.37&  0.36&  \textbf{0.27} & 0.32& 0.33\\
[0.8pt] \hdashline \noalign{\vskip 1pt}

\multirow{3}{*}{\makecell[l]{\ours \\ \textbf{(Ours)}}}
    &PSNR$\uparrow$&\textbf{22.33}&  \textbf{22.31} &  \textbf{20.55}& \textbf{24.59}& \textbf{22.44}\\
     &SSIM $\uparrow$&\textbf{0.78} &  \textbf{0.77} & \textbf{0.79} &  \textbf{0.85}& \textbf{0.80}\\
     &LPIPS $\downarrow$& \textbf{0.22}& \textbf{0.23} &0.29 & \textbf{0.18} & \textbf{0.23}\\ 
\bottomrule
\end{tabular}}
\caption{\textbf{Rendering Performance on EuRoC~\cite{euroc}.} Each scene result is the average of 3 runs. We take the numbers
for Photo-SLAM \cite{photo-slam} from their work. We successfully show the scalability of our system. Photorealistic 3D reconstruction comparison of large indoor environment EuRoC \cite{euroc} \texttt{MH-01} is shown in \cref{fig:teaser}}.
\label{tab:render_euroc}
\end{table}

\subsection{Ablations}

Post-processing, decay, and weighted depth loss are our system design choices. We present ablation studies to validate and support each of these design decisions.\\

 \begin{table}[h!]
  \centering
     \resizebox{\linewidth}{!}{
    \begin{tabular}{lccccc}
\toprule
       & GO-SLAM~\cite{go-slam} & GlORIE-SLAM~\cite{glorie} & MonoGS~\cite{gaussian-splatting-slam} &Splat-SLAM~\cite{splat-slam} & \textbf{Ours} \\
    \midrule

    GPU Usage [GiB] & 18.50  &  \nd 15.22 & \fst 14.62 &  17.57 & 16.20 \\[0.8pt] \hdashline \noalign{\vskip 1pt}
    Map Size [MB] & -  & 114.0 &  \nd 6.8 &  \fst 6.5 & 14.8 \\[0.8pt] \hdashline \noalign{\vskip 1pt}
    Avg. FPS &\nd 8.36  & 0.23 & 0.32 & 1.24 & \fst 9.94\\
    \bottomrule
    \end{tabular}}%
    \caption{Memory and Running Time Evaluation on Replica~\cite{Replica} \texttt{room0}. We measure the runtime statistics on the single process implementation of our method. We take the numbers from \cite{splat-slam} except for ours. Our peak memory usage and map size are comparable to existing works. Our method achieves to exhibit state-of-the-art 3D reconstruction in higher frame rates compared to other methods.} 
    \label{tab:memoryandruntime}
\end{table}

\noindent \textbf{Post Processing} We show post processing ablation results in \cref{tab:office0_refinement}. PSNR and Depth L1 metrics are recalculated for every 500 post-processing iterations. Our method exhibits a relatively small visual quality degradation when post-processing is skipped (indicated as 0K in \cref{tab:office0_refinement}) whereas visual quality significantly drops with no post-processing for Splat-SLAM \cite{splat-slam}. Our system exhibits diminishing returns with increased post-processing iterations. We attribute the fast convergence of our map and the minimal reliance on post-processing to our training strategy. \\

\begin{table}[h!]
    \def\dashline{\noalign{\vskip 3pt} \cdashline{2-11}\noalign{\vskip 3pt}}
    \centering
  \footnotesize
  \setlength{\tabcolsep}{0.7em}
  \resizebox{\linewidth}{!}{
    \begin{tabular}{lllcccc}
    \toprule
    Nbr of Final Iterations $\beta$ & Metric & \texttt{0K} & \texttt{0.5K} & \texttt{1K} & \texttt{2K}\\
    \midrule
    \multirow{2}{*}{\rotatebox{0}{\makecell[l]{Splat- \\SLAM~\cite{splat-slam}}}} 
      &PSNR $\uparrow$& 30.50 &	39.87&	40.59	& 41.20 \\
      &Depth L1 $\downarrow$ & 6.55	&2.37&	2.34&	2.40  \\

    \midrule
    \multirow{2}{*}{\rotatebox{0}{\makecell[l]{\textbf{Ours}}}}
      &PSNR $\uparrow$& \textbf{38.30}&	\textbf{40.92} &	\textbf{41.53}	&\textbf{41.68}\\
      &Depth L1 $\downarrow$ & \textbf{2.63}	&\textbf{2.18}&	\textbf{2.17}&	\textbf{2.30}\\

\bottomrule
\end{tabular}}
\caption{\textbf{Post-processing iterations ablation on Replica~\cite{Replica} \texttt{office0}}. The numbers for Splat-SLAM \cite{splat-slam} are taken from their work. Due to the fast convergence of mapping during tracking, we do not heavily rely on post-processing. The reconstruction benefits only a little from post-processing.} 
\label{tab:office0_refinement}
\end{table}

\begin{figure}[h!]
    \centering
    \includegraphics[width=\linewidth]{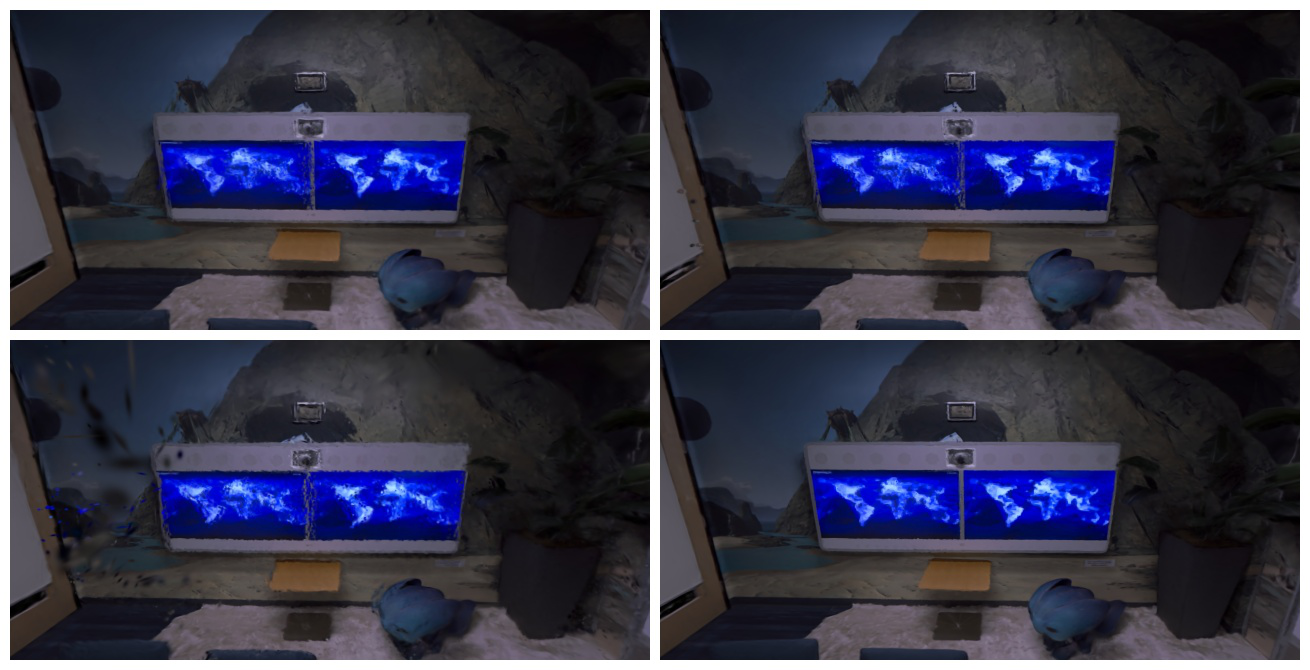}
    \caption{\textbf{Qualitative results for learning rate decay ablation study.} The four cases studied in \cref{tab:office0_decay} are shown in the figure. The results are given as constant learning rates of $\num{1.6e-4}$ at \textit{top-left}, \num{5e-5} at \textit{top-right}, $\num{1.6e-6}$ at \textit{bottom-left} and the decaying $\num{1.6e-4}$ learning rate at \textit{bottom-left} as reference.}
    \label{fig:decayab}
\end{figure}

\noindent \textbf{Decay} We demonstrate learning rate decay ablation in \cref{tab:office0_decay}. We compare 3 learning rates without decay with decaying learning rates. The selected 3 learning rates are $\text{lr}_f) =\num{1.6e-6}$ for lower bound,  $\text{lr}_i) =\num{1.6e-4}$ for upper bound, and the mean learning rate value $\num{5e-5}$ calculated according to \cref{eqn:decay}. We conduct this experiment with and without post-processing. As seen in no post-processing experiment in \cref{tab:office0_decay}, learning with decay greatly enhances the visual quality compared to other non-decaying learning rate setups. Qualitative results are shown in \cref{fig:decayab}. As observed, the fine details are not captured with non-decaying learning rates. Moreover, a post-processing step completely shadows the convergence problems of constant learning rate as seen in the experiment with post-processing in \cref{tab:office0_decay}. 

\begin{table}[h!]
    \def\dashline{\noalign{\vskip 3pt} \cdashline{2-11}\noalign{\vskip 3pt}}
    \centering
  \footnotesize
  \setlength{\tabcolsep}{0.7em}
  \resizebox{\linewidth}{!}{
    \begin{tabular}{lllcccc}
    \toprule
    Metric &Learning Rate  & \texttt{\num{1.6e-6}} & \texttt{\num{5e-5}} & \texttt{\num{1.6e-4}} & \texttt{\num{1.6e-4}} w/ decay \\
    
    \midrule
    \multicolumn{6}{l}{\cellcolor[HTML]{EEEEEE}{\textit{w/o Post Processing}}} \\ 
      PSNR $\uparrow$& & 31.92& 35.84 &34.71	& \textbf{38.30}\\
      Depth L1 $\downarrow$ & & 5.37	& 2.71& 2.76& \textbf{2.63}\\
    \midrule
     \multicolumn{6}{l}{\cellcolor[HTML]{EEEEEE}{\textit{w/ Post Processing}}} \\ 
      PSNR $\uparrow$& & 39.71& 39.91 &40.85	& \textbf{41.68}\\
      Depth L1 $\downarrow$ & & 2.73& \textbf{2.17}& 2.20& 2.30\\
\bottomrule
\end{tabular}}
\caption{\textbf{Learning Rate Hyperparameter Search on Replica~\cite{Replica} \texttt{office0}}. Our system benefits greatly from a slow learning rate combined with decay. In the presence of reliable depth maps, a high learning rate contributes to TV-static noise and slows down map convergence.} 
\label{tab:office0_decay}
\end{table}

\noindent \textbf{Depth Loss} The weighted depth loss ablation results are shown in \cref{tab:depth_loss_ablation}. The weighted depth loss that is given in \cref{eqn:depthloss} is compared to the scenarios with no depth loss in the overall loss function ($\alpha = 1$) and with raw depth values without weighting them by depth covariance. Post-processing is disabled to ensure the results are not obscured.

\begin{table}[h!]
\centering
\footnotesize
\setlength{\tabcolsep}{0.7em}

\begin{tabularx}{\linewidth}{XXXX}
\toprule
Metric  & Weighted& No Depth & Raw Depth\\
\midrule
PSNR$\uparrow$&\textbf{31.91} &31.56&30.81 \\
Depth L1 $\downarrow$&  \textbf{6.33}& 13.16 &  6.39 \\
\bottomrule
\end{tabularx}
\caption{\textbf{Weighted Depth Loss Ablation on Replica~\cite{Replica} \texttt{office2}.} Weighted depth loss enables better reconstruction without decreasing visual quality.}
\label{tab:depth_loss_ablation}
\end{table}

The weighted loss is superior to other choices as observed in \cref{tab:depth_loss_ablation}. A pure color loss performs well in terms of visual quality but deteriorates reconstruction quality. Using raw depth values in the loss function performs worse than the weighted loss regarding visual quality. Therefore, weighting the depth prevents visual quality from decreasing due to high uncertainty regions while keeping the reconstruction quality up by supervising depth. We speculate visual quality differences are not dramatic because our system initializes Gaussians according to depth maps regardless of the loss function. Therefore, initialized Gaussians are already in the vicinity of the corresponding depth value.

\section{Limitations}

The dense bundle adjustment is not feasible in full resolution. Therefore, dense depth maps are optimized at a lower resolution and upsampled back to the original resolution. We observe that this upsampling operation results in blurry edges. Therefore, utilizing upsampled dense depth maps to supervise the system results in poor performance at locations where sharp changes in depth occur.    

\section{Conclusion}

We showed that the depth supervision from a robust dense-SLAM method greatly enhances 3D reconstruction performance. Additionally, utilizing depth uncertainty as a mask for Gaussian initialization and as weights for depth loss aids the mapping process. We also highlighted the nuance between sparse and dense Gaussian initialization and its implications on mapping optimization. Our experiments demonstrated that dense SLAM-based 3D reconstruction can provide both state-of-the-art visual quality and a high frame rate even in relatively large scenes.

\clearpage

{  
    \small
    \bibliographystyle{ieeenat_fullname}
    \bibliography{main}
}
\clearpage
\setcounter{page}{1}
\maketitlesupplementary

\noindent IG-SLAM is a dense SLAM system capable of photorealistic 3D reconstruction, while simultaneously running at high frame rates. In this supplementary material, we provide additional results.

\section{Method}

We describe additional details about our method.

\subsection{Covariance Mask}

Assume the covariance for a depth map is given by $\Sigma$, we normalize covariance between [0,1] by \cref{eqn:norm_cov}
\begin{equation}
\Tilde{\Sigma}(u,v) = \frac{\Sigma(u,v) - \min{(\Sigma(u,v))}}{\max{(\Sigma(u,v))} - \min{(\Sigma(u,v))}}
\label{eqn:norm_cov}
\end{equation}

\noindent where $(u,v)$ are the pixel coordinates. A Maximum filter with a kernel size of 32 is applied to normalized covariance. Pixels with normalized covariance less than 0.2 are selected. Additionally, a majority filter with a kernel size of 32 is applied to obtain smooth valid regions in the mask.

\subsection{Pruning and Densification}

We follow the same procedure for pruning and identification in MonoGS \cite{gaussian-splatting-slam} n. Pruning is based on occlusion-aware visibility: if new Gaussians initialized in the last keyframes are not visible from this keyframe at the end of the optimization, they are removed. Additionally, for every 150 mapping iterations, Gaussians with opacity lower than 0.1 are removed. Densification is performed by splitting large Gaussians and cloning small ones in regions with high loss gradients, also every 150 mapping iterations.

\section{Additional Results}

We provide additional tracking and mapping results.

\section{Tracking}

We do not improve over GO-SLAM \cite{go-slam} in terms of tracking performance, as it is outside the scope of our work. However, we include the tracking results of Replica \cite{Replica}, TUM-RGB-D \cite{tumrgbd}, and ScanNet \cite{scannet} in \cref{tab:track_Replica}, \cref{tab:track_TUM}, and \cref{tab:track_scannet} for reference.

\begin{table}[htb]
    \centering
    \scriptsize
    \setlength{\tabcolsep}{5.4pt}
    \begin{tabular}{lcccccccccc}
    \toprule
    Metric  & \texttt{R-O} & \texttt{R-1} & \texttt{R-2} & \texttt{O-0} &\texttt{O-1} & \texttt{O-2} & \texttt{O-3} & \texttt{O-4} \\
    \midrule

    ATE(cm) &0.45 & 0.39 & 0.31 & 0.33 & 0.50 & 0.39 & 0.47 & 0.68  \\
    \bottomrule
    \end{tabular}
    \caption{
    \textbf{Tracking Accuracy ATE RMSE [cm] $\downarrow$ on Replica~\cite{Replica}.} Each scene result is the average of 3 runs.}
    \label{tab:track_Replica}
\end{table}

\begin{table}[htb]
    \centering
    \scriptsize
    \setlength{\tabcolsep}{5.4pt}
    \begin{tabular}{lcccccccccc}
    \toprule
    Metric  & \texttt{f1/desk} & \texttt{f2/xyz} & \texttt{f3/off} \\
    \midrule

    ATE(cm) & 2.73 & 0.35 & 2.08 \\
    \bottomrule
    \end{tabular}
    \caption{
    \textbf{Tracking Accuracy ATE RMSE [cm] $\downarrow$ on TUM-RGBD~\cite{tumrgbd}.} Each scene result is the average of 3 runs.}
    \label{tab:track_TUM}
\end{table}

\begin{table}[htb]
    \centering
    \scriptsize
    \setlength{\tabcolsep}{5.4pt}
    \begin{tabular}{lcccccccccc}
    \toprule
    Metric  & \texttt{0000} & \texttt{0059} & \texttt{0106} & \texttt{0169}& \texttt{0181} & \texttt{0207} \\
    \midrule

    ATE(cm) & 6.16 & 71.46 & 7.38 & 8.46 & 8.60 & 9.55 \\
    \bottomrule
    \end{tabular}
    \caption{
    \textbf{Tracking Accuracy ATE RMSE [cm] $\downarrow$ on ScanNet~\cite{scannet}.} Each scene result is the average of 3 runs.}
    \label{tab:track_scannet}
\end{table}

\subsection{Mapping}

\begin{figure*}[ht!] 
  \centering
  \includegraphics[width=\textwidth]{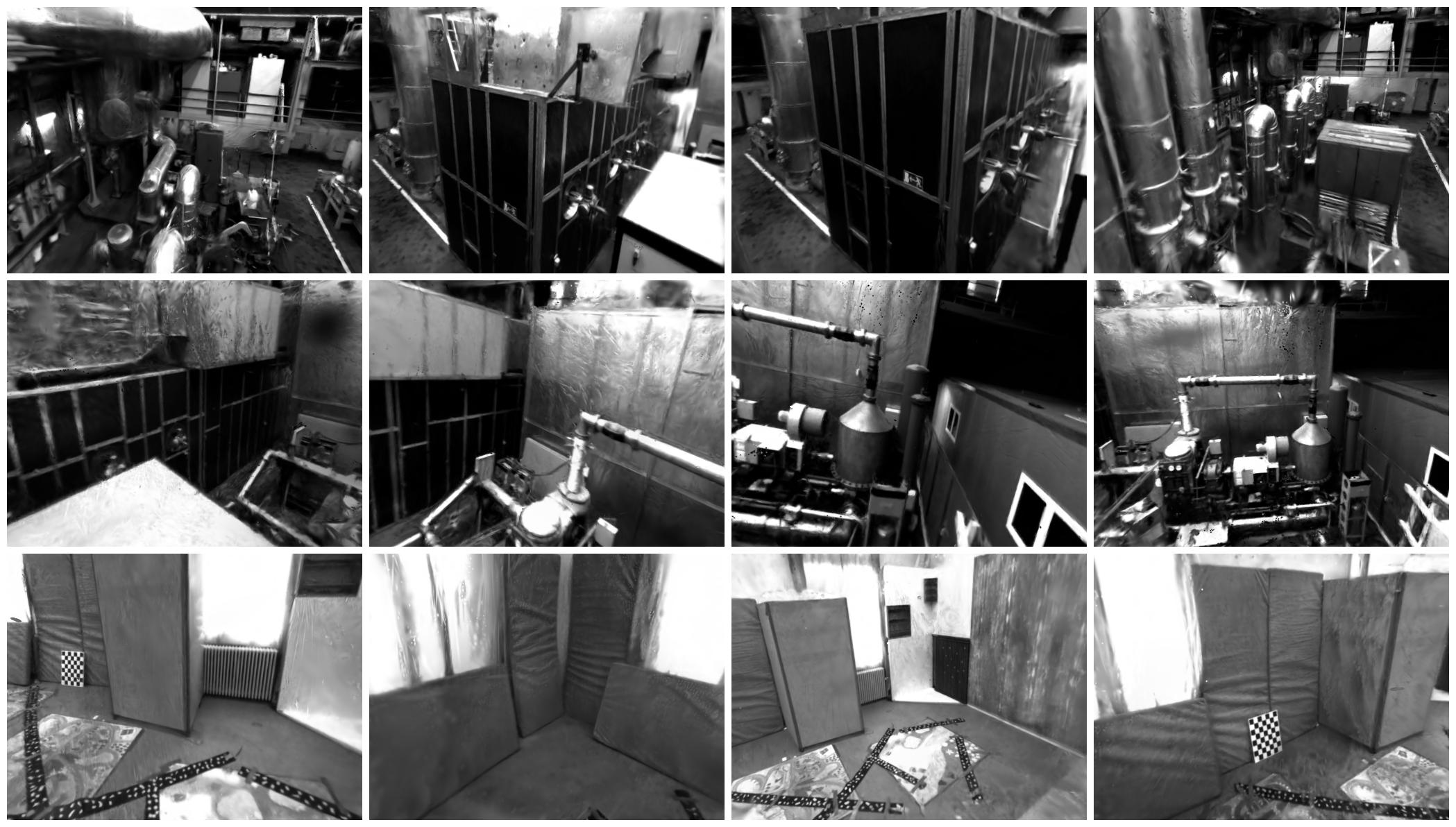} 
  \caption{\textbf{Qualitative results of IG-SLAM on EuRoC \cite{euroc}.} The results in the \textit{top row}, \textit{middle row}, and \textit{bottom row} are from \texttt{MH-02}, \texttt{MH-03}, \texttt{V1-01} respectively.}
  \label{fig:quali}
\end{figure*}

The results of each scene of the Replica \cite{Replica} are given in \cref{tab:replica_full}. Full evaluations on EuRoC \cite{euroc} Machine Hall and Vicon Room are given in \cref{tab:euroc_full_mh} and \cref{tab:euroc_full_v}. Moreover, additional qualitative results of EuRoC \cite{euroc} are exhibited in \cref{fig:quali} 

 \begin{table}[h!]
\centering
\footnotesize
\setlength{\tabcolsep}{0.7em}
\resizebox{\linewidth}{!}{
\begin{tabular}{llccccccc}
\toprule
Metric & \texttt{R-0} & \texttt{R-1} & \texttt{R-2} & \texttt{O-0} & \texttt{O-1} & \texttt{O-2} & \texttt{O-3} & \texttt{O-4} \\
\midrule

PSNR$\uparrow$  & 32.33 & 34.64 & 35.29 & 41.68 & 41.30 & 34.68 & 34.92 & 34.80 \\
SSIM $\uparrow$ &0.93 & 0.95 &  0.96 & 0.98 & 0.98 & 0.95 & 0.96 & 0.96 \\
LPIPS$\downarrow$ & 0.07 & 0.06 & 0.05 & 0.02 & 0.03 & 0.06 & 0.05 & 0.07 \\
\makecell[l]{Depth L1}$\downarrow$ & 4.79 & 3.04 & 4.15 & 2.23 & 1.94 & 6.40 &7.67 & 4.45\\

\bottomrule
\end{tabular}}

\caption{\textbf{Full evaluation on Replica~\cite{Replica}.} Each scene result is the average of 3 runs.}
\label{tab:replica_full}
\end{table}

\begin{table}[h!]
\centering
\footnotesize
\setlength{\tabcolsep}{0.7em}
\resizebox{\linewidth}{!}{
\begin{tabular}{llcccccccccc}
\toprule
Metric & \texttt{MH-01} & \texttt{MH-02} & \texttt{MH-03} & \texttt{MH-04} & \texttt{MH-05}\\
\midrule

PSNR$\uparrow$ & 22.33 & 22.31 & 20.78 & 23.62 & 19.85 \\
SSIM $\uparrow$ & 0.78 & 0.77 & 0.71 & 0.82 & 0.70 \\
LPIPS$\downarrow$ & 0.22 & 0.23 & 0.28 & 0.19 & 0.35 \\

\bottomrule
\end{tabular}}

\caption{\textbf{Full evaluation on EuRoC~\cite{euroc} \texttt{Machine Hall}.} Each scene result is the average of 3 runs.}
\label{tab:euroc_full_mh}
\end{table}

\begin{table}[h!]
\centering
\footnotesize
\setlength{\tabcolsep}{0.7em}
\resizebox{\linewidth}{!}{
\begin{tabular}{llcccccccccc}
\toprule
Metric  & \texttt{V1-01} & \texttt{V1-02} & \texttt{V1-03} & \texttt{V2-01} & \texttt{V2-02} & \texttt{V2-03} \\
\midrule

PSNR$\uparrow$ & 20.55 & 22.86 &20.11 & 24.59 & 23.70& 21.62\\
SSIM $\uparrow$& 0.79  & 0.84 &0.74 & 0.85 & 0.83 &0.74\\
LPIPS$\downarrow$ & 0.29& 0.26 &0.42 & 0.18 & 0.23 & 0.41\\

\bottomrule
\end{tabular}}

\caption{\textbf{Full evaluation on EuRoC~\cite{euroc} \texttt{Vicon Room}} Each scene result is the average of 3 runs.}
\label{tab:euroc_full_v}
\end{table}

\end{document}